\documentclass[lettersize,journal]{IEEEtran}

\usepackage{graphicx}
\usepackage{xcolor}
\usepackage[table]{xcolor}
\usepackage{amsmath,amsfonts,amssymb}
\usepackage{booktabs} 
\usepackage{multirow} 
\usepackage{tcolorbox}
\usepackage{caption}
\usepackage{subcaption} 
\usepackage{pdfpages}
\usepackage{url}
\usepackage{adjustbox}
\usepackage{bm}

\title{\LARGE \bf
Deformable Cluster Manipulation via Whole-Arm Policy Learning
}

\author{Jayadeep Jacob$^{1,2}$, Wenzheng Zhang$^{1}$, Houston Warren$^{1}$,
 \\ 
Paulo Borges$^{3}$, Tirthankar Bandyopadhyay$^{2}$, Fabio Ramos$^{1,4}$

\thanks{$^{1,2}$Jayadeep Jacob, $^{1}$Wenzheng Zhang, and $^{1}$Houston Warren are with the School of Computer Science, The University of Sydney, NSW, Australia. Jayadeep Jacob is also with Data61, CSIRO, Pullenvale, QLD, Australia.
        {\tt\footnotesize jjac4485@sydney.edu.au}; {\tt\footnotesize jay.jacob@data61.csiro.au}; {\tt\footnotesize \{wzha2981, houston.warren\}@sydney.edu.au}}%

\thanks{$^{2}$Tirthankar Bandyopadhyay is with the CyberPhysical Systems Program, Data61, CSIRO, Pullenvale, QLD, Australia. 
        {\tt\footnotesize tirtha.bandy@csiro.au}}%

\thanks{$^{3}$Paulo Borges is with Orica. 
        {\tt\footnotesize paulo.borges@orica.com}}%

\thanks{$^{1,4}$Fabio Ramos is with the School of Computer Science, The University of Sydney, NSW, Australia, and with the NVIDIA Corporation, Seattle, USA
        {\tt\footnotesize fabio.ramos@sydney.edu.au}}

}
\begin{document}

\maketitle
\thispagestyle{empty}
\pagestyle{empty}

\begin{abstract}

Manipulating clusters of deformable objects presents a substantial challenge with widespread applicability, but requires contact-rich whole-arm interactions. A potential solution must address the limited capacity for realistic model synthesis, high uncertainty in perception, and the lack of efficient spatial abstractions, among others. We propose a novel framework for learning model-free policies integrating two modalities: 3D point clouds and proprioceptive touch indicators, emphasising manipulation with full body contact awareness, going beyond traditional end-effector modes. Our reinforcement learning framework leverages a distributional state representation, aided by kernel mean embeddings, to achieve improved training efficiency and real-time inference. Furthermore, we propose a novel context-agnostic occlusion heuristic to clear deformables from a target region for exposure tasks. We deploy the framework in a power line clearance scenario and observe that the agent generates creative strategies leveraging multiple arm links for de-occlusion. Finally, we perform zero-shot sim-to-real policy transfer, allowing the arm to clear real branches with unknown occlusion patterns, unseen topology, and uncertain dynamics.\newline Website: \textcolor{blue}{\underline{\url{https://sites.google.com/view/dcmwap/}}}
\end{abstract}

\section{INTRODUCTION}

Learning to manipulate deformable objects poses a formidable challenge in robotics \cite{yin2021modeling}; however, operating with clusters of deformables, such as cable bundles, multi-branch tree canopies, and cloth piles, is significantly harder. Although imprecise physics knowledge \cite{arriola2020modeling} and sensor uncertainties are contributing factors, the principal challenge lies in inferring a compact, low-dimensional state abstraction and the temporal state evolution, based on noisy perception from a high-dimensional ground truth \cite{antonova2022bayesian}. Furthermore, with deformable clusters, tiny differences in contact locations, material properties, and initial configuration among the individual cluster constituents (e.g. between thicker and thinner plant stems) give rise to non-linear resistance forces and divergent motion trajectories upon interaction. The resulting chaotic dynamics render model-based learning infeasible, while model-free solutions have high sample complexity, and therefore, the compactness and generation efficiency of the state representation gather significance \cite{maillard2011selecting}.

\begin{figure}[!t]
    \centering
    \includegraphics[width=0.49\textwidth]{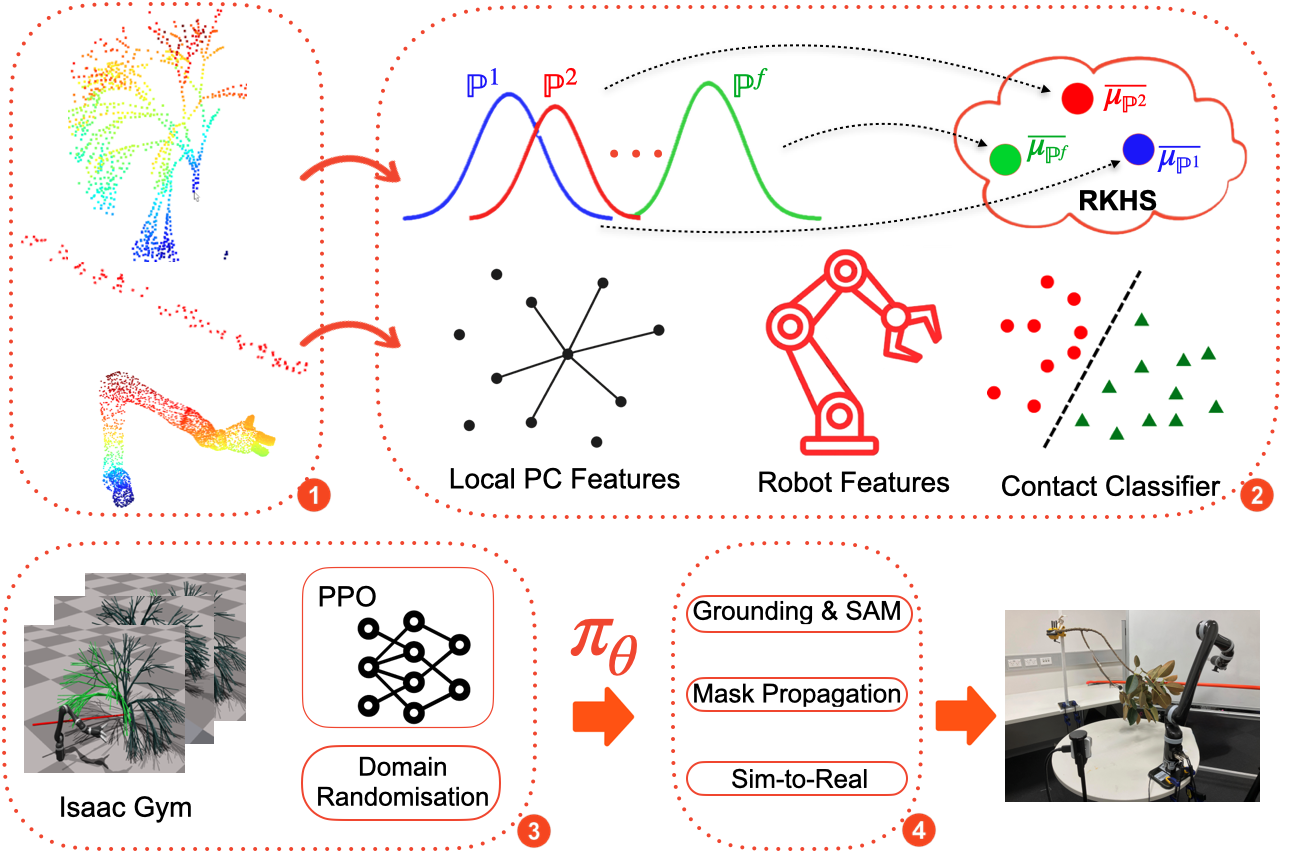}
    \caption{Overview: (1) Segmented point clouds corresponding to the clustered deformable branches, clearance region, and the robot are captured from the scene. (2) Distribution embeddings representing global scene are generated via kernel mean operator. Additional features include neighbourhood point features, robot sensor metrics, and proprioceptive contact indicators. (3) RL training with domain randomised geometry and dynamics on Isaac Gym parallel simulator. (4) Inference \& zero-shot transfer to the real world aided by Grounding DINO, SAM-HQ, and Cutie frameworks.}
    
    \label{fig:block_diagram}
    \vspace*{-4mm}
\end{figure}

Novel deformable works manipulating cables \cite{grannen2020untangling}, dough \cite{lin2022planning}, and clothes \cite{deng2024general} are all planning approaches; furthermore, they exclusively focus on end-effector-based control. In a cluster context, such grasp-based solutions tailored to individual deformable members become prohibitively long. On the contrary, a multi-link manipulation strategy substantially escalates the contact complexity, necessitating additional perception modalities apart from vision. However, typical auxiliary sensors for whole-arm contact detection are either expensive \cite{jain2013reaching} or too slow \cite{huang20243d} for low-latency applications.

Addressing these challenges, we propose a model-free, non-prehensile, multi-modal, reinforcement learning (RL) approach operating on the cluster as a whole with full-body contact, but without external tactile sensing. We use a distributional representation \cite{antonova2022bayesian} of the cluster state generated by embedding point cloud into a Reproducing Kernel Hilbert Space (RKHS), without key-point intermediaries. Compared to traditional point cloud feature extractors, a distribution embedding is notably lightweight, enabling faster training and real-time inference. It eliminates the need for points to track identical cluster locations, is invariant to permutations, and can handle variable input sizes due to members moving out of camera view between frames. Further, the observations include a whole-arm contact classifier \cite{jacobgentle} output from proprioceptive input to flag contacts. Finally, we address the high sample requirements of on-policy RL by leveraging the parallelism of physics simulators during training and transferring it zero-shot to the real-world.

\begin{figure} 
\begin{tabular}{cc}
    \includegraphics[height=0.16\textwidth]{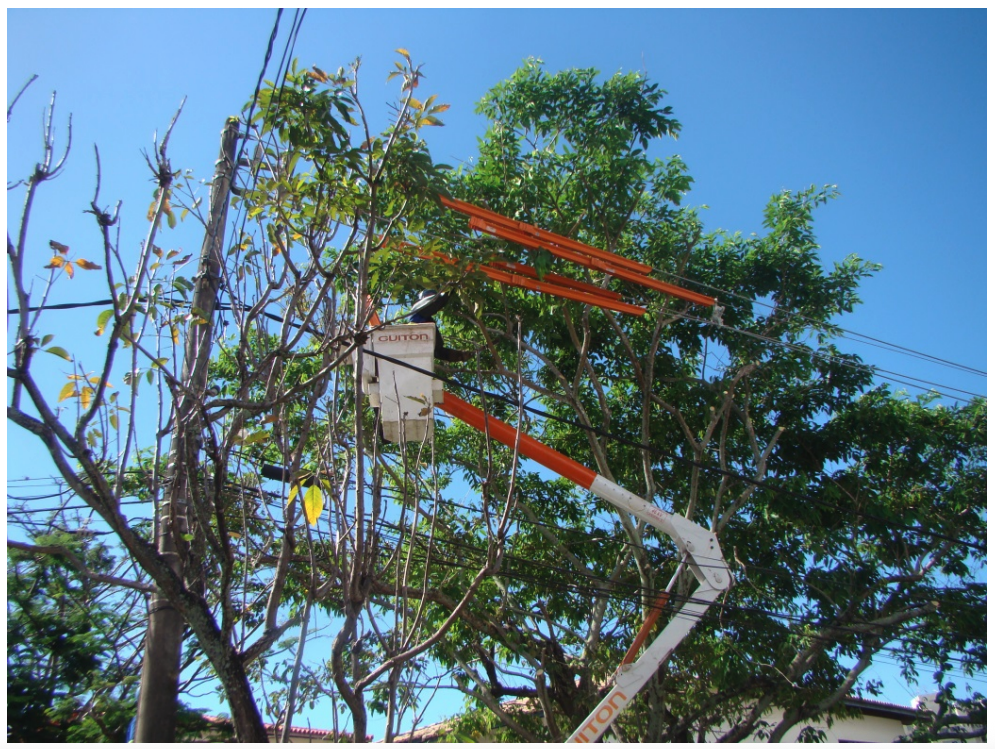}&
    \includegraphics[height=0.16\textwidth] {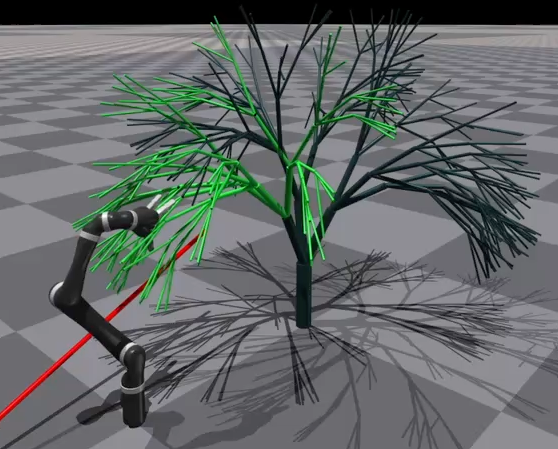} \\
     (a) & (b)

\end{tabular}

\caption{ \textcolor{black}{(a) Manual pruning of overhanging branches impeding power lines, from \cite{siebert2014survey} (b) Our simulation setup with an L-system structure for the power line clearance scenario.}}
\vspace*{-3mm}
\label{fig:problem-description}
\end{figure}

This work focuses on a specific subclass of problems, aiming to de-occlude local regions of the deformable cluster; for instance, to clear a cylinder-shaped region among tree branches, by manipulating a few of them together. \textcolor{black}{We implement two illustrative applications of such a skill abstraction}:

\begin{enumerate}
    \item \textbf{Clearing Power Lines}: Hazardous tree branches encroaching on overhead power infrastructure is a leading cause of forest fires and outages \cite{lowe2022identification}\cite{guggenmoos2003effects}. Pruning foliage near overhead lines (Fig. \ref {fig:problem-description}a).  is dangerous for humans due to risks like falls, shocks, and tool accidents \cite{siebert2014survey}; therefore, robotic assistive systems that can clear the branches away from the lines before pruning can enhance safety while reducing outages. 
    \item \textbf{Agricultural Exposure}:  Modern harvesting systems suffer from lower success rates with partially (50\%-75\%) and fully occluded (5\%) fruits \cite{zhou2022intelligent}.
    Manipulating plant stems for exposing fruits or a diseased region to an external inspection camera requires a line-of-sight restoration through multi-branch clearance.
\end{enumerate}

Among these, the first problem presents the greatest challenge due to additional constraints imposed by the rigid power line on robot's motion, and thus serves as the primary focus of this paper in both simulation and real settings. While our aim is not to provide an industry-grade system, it is notable that no autonomous solutions currently exist for the power line problem. \textcolor{black}{Additionally, we provide simulation videos for the agricultural exposure task to emphasise the generality of our method.} Specifically, our contributions are:

\begin{enumerate}
\color{black}
\item A multi-modal, whole-body contact policy framework to manipulate deformable clusters, learned from both point clouds and proprioceptive touch detection inputs.
\item An efficient distributional state representation of the complex deformable cluster geometry, with low computational overheads, in an RL policy learning context.

\item A novel context-agnostic occlusion heuristic and reward strategy generalisable to various exposure applications to clear deformables from a target region.

\item A specific implementation to clear power lines of overhanging tree branches and zero-shot sim-to-real transfer that handles unseen branch topology and displays novel clearance patterns in simulation and real.

\end{enumerate}

\textcolor{black}{Our first two contributions are technical and apply to deformable manipulation in general, while the third and fourth are system-level, focusing on de-occlusion, but deployable beyond trees.} Subsequent sections specifically focus on the power line clearance problem; however, the methodology is readily adaptable to others. We encourage readers to view the supplementary videos to see our approach in action.

\section{RELATED WORK}

For deformable manipulation with high-dimensional geometries, point clouds offer a rich sensor-agnostic scene mapping. Notable 1D  implementations for tracking \cite{tang2022track} and learning dense depth object descriptor features \cite{sundaresan2020learning} leverage point representations. Point clouds have been used to learn policies with 2D and 3D deformables as well for cloth manipulation \cite{deng2024general}, liquid scooping \cite{seita2023toolflownet}, and dough cutting \cite{lin2022planning}. On the policy front, these deformable works use either Motion Planning with primitives \cite{deng2024general}, Task and Motion Planning \cite{lin2022planning}, or Behavioural Cloning \cite{sundaresan2020learning}. In contrast, model-free RL works on deformables like ours are scarce, largely due to poor sample efficiency. With rigid body point clouds, model-free policies have been proposed \cite{liu2022frame} for interactive tasks, including sim-to-real transfer. 

Post-scene capture, deformable representation strategies include keypoints \cite{tang2022track}, graphs \cite{ma2021learning}, and dense-descriptors \cite{sundaresan2020learning}. 
%
%
Relevant to our work, the deformable state can be represented as as a distributional embedding with a kernel mean operator \cite{antonova2022bayesian}.
They tackle a real-to-sim inference problem with an RKHS-net layer using a small number of keypoints ($\approx 10$) extracted from images. In contrast, we project entire point clouds ($\approx 512$) of multiple objects to RKHS and use the embeddings directly as RL observations in a sim-to-real context. Our policy enables the arm to contort itself into hooks and props, leveraging the whole arm for clearance. While such formations are less prevalent, contact-aware reaching has been achieved with external full-body tactile sensors, for rigid clutter \cite{jain2013reaching}, or with internal sensors for tree branches \cite{jacobgentle}. However, we go beyond simple reach to sophisticated clearance tasks and implement a multi-modal policy integrating touch and vision feedback.

Among deformables, plant interaction is substantially harder owing to the non-uniformity in geometry, dynamics, and visual features. While \cite{yao2024safe} gently moves a leaf aside to estimate hidden fruit pose, \cite{zhang2023push} clears multiple leaves simultaneously with the end-effector using large quantities of real-world data. Both works assume soft leafy resistance, which is typically addressed in farms with simpler tricks, such as blowing compressed air \cite{yamamoto2014development} to separate leaves and fruits. From a learning perspective, \cite{yandun2021reaching} and \cite{subedi2025find} employ vision-based RL for reaching vine pruning locations and for exposing fruits, respectively; however, both assume known plant topology while focusing on the end-effector. In comparison, our approach accommodates stiff resistance from multiple branches, uses the whole arm to maximise applied torques, and is trained in simulation without digital twins, plant mesh similarity, or real-world data. Finally, none of the aforementioned works are designed for the hybrid rigid-deformable problem of power line clearance. 

\section{APPROACH}

\textcolor{black}{A brief overview of our approach (Fig. \ref{fig:block_diagram}) is as follows. First, we generate parallel simulations of deformable tree-branches (Sec. \ref{sec:deformable_simulation}) to train an RL policy (Sec. \ref{sec:rl_policy_learning}). To guide policy learning, we compute an occlusion metric from the scene, which serves as a reward signal (Sec. \ref{sec:occlusion_heuristic_reward}). The policy is trained in a physics simulator with domain randomisation applied to the geometry and dynamics of trees (Sec. \ref{sec:domain_randomisation_for_rl}). The learning framework leverages a kernel mean operator (Sec. \ref{sec:kernel_mean_embedding}) to encode the segmented point clouds as RL features (Sec. \ref{sec:distribution_embedding_for_deformables}). In addition, the features include robot metrics and the output of a touch classifier. 
For sim-to-real transfer (Sec. \ref{sec:hardware_and_sim_to_real}), we use several vision-based segmentation and masking methods to process camera images, generate object masks and extract the vision features necessary for real-world policy deployment (Sec. \ref{sec:real_vision_pipeline}). 
}

\subsection{Deformable Simulation}
\label{sec:deformable_simulation}
Realistic simulation of the intricate tree-branch geometry is challenging; fortunately, this is well-studied in literature under the L-system paradigm \cite{prusinkiewicz2012algorithmic}, predominantly for visualisation. An L-system borrows from formal grammar theory to model plant morphology by applying a recursive rule collection, starting from an axiom, mimicking the branching patterns. Recently, \cite{jacobgentle} has shown that this paradigm can be exported to 3D physics simulators, by representing branch segments as rigid cylindrical links connected by revolute joints, amenable to robotic interaction. In this setup, branch deformations are modelled with a mass-spring-damper \cite{jacob2024learning} actuated with proportional-derivative controllers. We borrow the recursion rules, growth
attributes, and L-system parameters from \cite{jacobgentle}; however, we extend it with spherical joints to enable multi-axis deformation and enhance compliance, refer Fig. \ref {fig:problem-description}(b). We run thousands of such tree models in parallel on the distributed physics simulator Isaac
Gym \cite{makoviychuk2021isaac}.

\subsection{Policy Learning}
\label{sec:rl_policy_learning}
We formulate the de-occlusion task as a discrete-time Markov Decision Process (MDP), where an agent learns a stochastic policy $\pi_{w}(a|o)$; parameterised by weights $w$. An MDP is defined by $\langle S, A, P_{a}, R_{a}, \beta \rangle$, where $S$ is states, $A$ is actions, $P_{a}$ and $R_{a}$ are the state transition model and the reward from action $a \in A $, and $\beta$ is a discount factor. The agent aims to maximise $\mathbb{E}_{\pi} \left[ \sum_{t=0}^{T-1} \beta^t R_{a}(s_t, a_t) \right]$, the expected discounted rewards over $T$ steps. Crucially, the agent only has access to noisy observations $o \in O$; for instance, point clouds from the camera or robot sensor measurements rather than the true state. Our RL algorithm choice is Proximal Policy Optimisation \cite{schulman2017proximal}.

\subsection{\textcolor{black}{Occlusion Heuristic as Reward Signal}}
\label{sec:occlusion_heuristic_reward}

A key consideration for de-occlusion is determining an effective yet simple measure to capture the temporal progression in occlusion level from manipulation, \textcolor{black}{for instance, as a reward signal to guide the policy learning}. Common metrics include visible surface ratio, pixel counts \cite{bai2025retrieval} or labelled occlusion levels. However, these metrics depend either on prior knowledge of the target geometry or partial observability at start, and may not be smooth enough for an RL agent. In contrast, our occlusion heuristic $h_t \in [0, 1]$ is independent of target characteristics or its presence.

Given segmented point clouds $P^{(1)}  \in \mathbb{R}^{N_1 \times 3}$, $P^{(2)}  \in \mathbb{R}^{N_2 \times 3}$ of two entangled entities, the obstructing branches $P^{zbr}$ and a virtual clearance region $P^{clr}$ in our case, the $h_t$ is computed as follows. First, we define a pairwise distance matrix \( D \in \mathbb{R}^{N_1 \times N_2} \): 
\[
D_{i,j} = \| p^{(1)}_i - p^{(2)}_j \|_2 \text{ for } i = 1, \dots, N_1; \ j = 1, \dots, N_2.
\]
Let \( \mathcal{S} \subset \{1, \dots, N_1\} \times \{1, \dots, N_2\} \) be the indices set corresponding to the \( k \) smallest distances in \( D \), i.e., the $(i,j)$ indices of the \( k \) nearest-neighbor point pairs, and $d_{th}$ be a  distance threshold, then $\mathcal{D}_k := \left\{ D_{i,j} \mid (i,j) \in \mathcal{S} \right\}$ and

\begin{equation}
h(P^{(1)}, P^{(2)}) := \frac{1}{k} \sum_{d \in \mathcal{D}_k} \mathbb{I}(d < d_{\text{th}}).
\end{equation}

Intuitively, it is the fraction of point pairs (from different groups) that breach a proximity threshold to the total number of pairs in the neighbourhood, where both the threshold $d_{th}$ and the neighbourhood size $k$ are tunable. Larger the number of branch segments covering the clearance region, higher the $h_t$. \textcolor{black}{The threshold $d_{th}$ (set to 10cm) is the safe distance for an auxiliary robot to prune the cleared branches. The $k$ (set to 200) balances the branch-line interaction fidelity against computational cost; a lower $k$ risks skipping critical contact points, while a higher $k$ significantly increases processing time.}
We emphasise the clearance region has no shape constraints due to its point representation;  for instance, it can be defined to de-occlude an expected curved end-effector trajectory of a second manipulator in a dual-arm setting. 

\subsection{\textcolor{black}{Domain Randomisation for RL training}}
\label{sec:domain_randomisation_for_rl}
This work aims to generalise a learned policy to deformables with unseen geometry, undetermined dynamics, and devoid of chromatic affinity, by not relying on digital twins \cite{subedi2025find}, parameter inference \cite{jacob2024learning}\cite{antonova2022bayesian}, or RGB images during training. Instead, we perturb simulation parameters, for example, L-system traits such as branch divergence angle and elongation rate, and dynamics parameters such as spring stiffness and damping. Further, we subsample and permute all point clouds while adding noise at each time-step. 

\subsection{\textcolor{black}{Kernel Mean Embedding for Observation Encoding}}
\label{sec:kernel_mean_embedding}

In this section, we briefly examine distribution embedding and its spectral approximation; for a thorough explanation, see \cite{muandet2017kernel}\cite{rahimi2007random}. An RKHS is a Hilbert space of functions uniquely determined by a positive definite kernel function $k : \mathcal{X} \times \mathcal{X} \to \mathbb{R}$, that guarantees both an implicit feature mapping $\varphi : \mathcal{X} \to \mathcal{H}$ from the input space and the higher-dimensional target space $\mathcal{H}$. Akin to point mappings, a kernel mean operator lifts a probability distribution $\mathbb{P}$ to a single mean function in $\mathcal{H}$, however, without any loss of information for a characteristic kernel such as RBF: $k(x,x_{}^{\prime}) = \exp(-\frac{||x- x_{}^{\prime}||_{}^{2}}{2\gamma{}^{2}})$. We imagine $x$ to be points from the 3D point clouds. The lifted kernel mean is defined as: 
\begin{equation}
\label{eq:kme}
    \varphi(\mathbb{P}) = \mu_{\mathbb{P}} :=  \mathbb{E}_{x \sim \mathbb{P}}[k(\cdot, x)] = \int k(\cdot, x) \, d\mathbb{P}
\end{equation}
Intuitively, formulation (\ref{eq:kme}) could be viewed as the average kernel similarity of parameter $x$ to all domain points according to the distribution $\mathbb{P}$ in RKHS. When the underlying distribution is not explicitly known, which often is the case, an empirical kernel mean $\overline{\mu_{\mathbb{P}}}$ can approximate the true $\mu_{\mathbb{P}}$ from just the i.i.d samples $\{x_1, \dots, x_N\}$ drawn from $\mathbb{P}$. 
\begin{equation}
\label{eq:empirical_kme}
\overline{\mu_{\mathbb{P}}} := \frac{1}{N} \sum_{i=1}^{N} \varphi(x_i) = \frac{1}{N^2} \sum_{i=1}^{N} \sum_{j=1}^{N} \langle \varphi(x_i), \varphi(x_j) \rangle_{\mathcal{H}}
\end{equation}
While the inner product $\langle .,. \rangle_{\mathcal{H}}$ in eqn. (\ref{eq:empirical_kme}) can indeed be computed with a finite-dimensional gram matrix, leveraging the reproducing property $\langle \varphi(x_i), \varphi(x_j) \rangle_{\mathcal{H}} = k(x_i, x_j)$, but this is still problematic for large N. On the other hand, representing kernel functions through their spectral representations, particularly with Random Fourier Features (RFF) \cite{rahimi2007random}, significantly reduces the computational overhead. Through Bochner's theorem, any kernel can be defined as a Fourier transform of a non-negative measure, subject to a stationarity constraint $k(x,x') = k(x-x')$.  

A Monte Carlo estimate of the kernel can then be written as $k(x, x') \approx \frac{1}{R}\sum_{r=1}^R \cos({\omega}_r^\top(x-x'))$ where the frequencies $\omega_r$ are sampled from a spectral density, $\omega_r \sim p(\omega)$. The above estimate becomes computationally efficient for $R<<N$, which we exploit. If $k(x, x') \approx \hat{\varphi}^\top(x)\hat{\varphi}(x') $, the empirical kernel embedding can be further simplified as:
\begin{equation*}
\overline{\mu_{\mathbb{P}}} := \frac{1}{N} \sum_{i=1}^{N} \frac{1}{\sqrt{R}} \begin{bmatrix}
\cos(\omega_1^\top x_i), \sin(\omega_1^\top x_i), \dots, \sin(\omega_R^\top x_i)
\end{bmatrix}.
\end{equation*}
\textcolor{black}{A higher $R$ indicate more informative features, but an excessive value explodes the observation space, demanding more samples during training. As a trade-off, we choose $R=16$ and use the RBF kernel, which results from the Fourier transform of a Gaussian $p(\omega) \sim \mathcal{N}(0,1)$.}

\subsection{\textcolor{black}{Distribution Embedding for Deformable Clearance}}
\label{sec:distribution_embedding_for_deformables}

\textcolor{black}{Given a point cloud $P_t \in \mathbb{R}^{N \times 3}$ at time-step $t$, we assume it as samples from an underlying distribution. These 3D points are elevated to RKHS with an RFF approximated kernel mean to form the key constituent of RL observations.} We capture four independent point clouds of varying sizes: $P^{rob}$, $P^{clr}$, $P^{wbr}$, $P^{zbr}$, representing the robot, clearance region (i.e. a slice of the power line), the whole tree, and a zoomed-in version of the branches near the occlusion. While the robot points are sampled from URDF mesh based on link positions at each step, the cylindrical clearance region points are sampled from its surface, given a fixed pose. In contrast, camera sensors provide the deformable and time-varying points of the occluding branches.

There are several advantages to such a representation. Firstly, although the explicitly mapped feature locations are unknown, the agent can exploit the relative similarity of embedded points in RKHS, for instance, to minimise the foliage-power line entanglement or maximise the gripper-branch proximity, resembling the `kernel trick' common in machine learning. Secondly, a distributional interpretation is invariant to point permutations and robust to the high noise content characteristic of streaming point clouds. Third, unlike conventional point cloud feature extractors like PointNet++ \cite{qi2017pointnet++}, kernel embeddings are resilient to variations in the size and structure of input point clouds, amplified by the arm, branches and the power line frequently occluding each other. Finally, and most crucially, this approach is significantly faster, reducing training time and enabling real-time inference, as our experiments demonstrate. This speed advantage stems from formulation in Section \ref{sec:kernel_mean_embedding}, where the feature map $\hat{\varphi}(x)$ can be derived with a single matrix product of the input points $x$ and the pre-computed Fourier coefficients ${\omega}_r^\top$ of complexity $\mathcal{O}(N \times R)$, followed by a sin/cos transformation, and an additional row-average to obtain the embedding $\overline{\mu_{\mathbb{P}}}$, a highly efficient pipeline scalable to thousands of environments, amenable to GPU parallelisation.

\subsection{Observation \& Reward Space}

\begin{table}[ht]
\centering
\caption{Observation Feature Groups}
\label{tab:rl_obs_features}
\begin{tabular}{llr}
\toprule
\textbf{Group} & \textbf{Feature} & \textbf{Shape} \\
\midrule
\multirow{3}{*}{Proprioceptive}
    & joint pose: ${q}$ & 6 \\
    & joint velocity: $\dot{{q}}$ & 6 \\
    & ee quaternion: ${r}_{\text{EE}}$ & 4 \\
\midrule
\multirow{3}{*}{KME} 
    & whole branches: $\overline{\mu_{\mathbb{P}}}[P^{wbr}]$  & 16 \\
    & zoomed-in branches: $\overline{\mu_{\mathbb{P}}}[P^{zbr}]$  & 16 \\
    & clearance region: $\overline{\mu_{\mathbb{P}}}[P^{clr}]$  & 16 \\
    & robot: $\overline{\mu_{\mathbb{P}}}[P^{rob}]$  & 16 \\
\midrule
\multirow{1}{*}{Touch} 
    & contact indicator y/n: $\mathbb{I}(\Vert F_t \Vert_2 > f_u)$ & 1 \\
\midrule
\multirow{2}{*}{Local PC} 
    & ee-branch dist: $\mathcal{D}_{k=5}(P^{ee}, P^{wbr})$ & 5 \\
    & safety breach indicator y/n & 1 \\
\midrule
\multirow{1}{*}{Occ Heuristic} 
    & $h_t(P^{wbr}, P^{clr})$ & 1 \\
\midrule
\multicolumn{2}{l}{\textbf{All Features}} & \textbf{88} \\
\bottomrule
\end{tabular}
\vspace*{-3mm}
\end{table}

Our observation space $O$ consists of five feature sets, listed in Table \ref{tab:rl_obs_features}. First, we construct two feature sets from point clouds: the KME features described in Section \ref{sec:distribution_embedding_for_deformables}, which characterise global structure, and a local feature group, which focuses on nearby point distances. The latter includes the Euclidean distance $d$ of the end-effector to the closest $k=5$ branch points and a safety breach indicator, computed as: 
\[
\delta_{sm} = \mathbb{I} \left\{ \frac{1}{|\mathcal{D}_5|} \sum_{d \in \mathcal{D}_5} d < d_{\text{sm}} \right\} ;\mathcal{D}_5 = \mathcal{D}_{k=5}(P^{rob}, P^{clr}).
\]
\textcolor{black}{A robot-power line contact doesn't compromise the validity of our approach, as the overhanging branches act as conductors already\cite{guggenmoos2003effects}; nevertheless, a safety margin ($d_{sm}=4$cm, to account for motion and sensor uncertainties) is introduced to reduce power line wear and tear.}

A novel aspect of our method is its multi-modality, attributable to a whole-arm touch detector. Recently, \cite{jacobgentle} has shown that in the absence of expensive tactile sensors, time-series sliding window torque measurements: $[\tau_t, \tau_{t-1}, \tau_{t-2},...,\tau_{t-(m-1)}]$ over the last $m$ time steps along with other proprioceptive features, can be used to build a classifier trained to detect contact `bumps' at the current time-step $t$, from labeled touch data. While \cite{jacobgentle} is trained with touch features for a simple reach task, we extend that approach and demonstrate that the touch classifier is useful even in the presence of vision and can aid complex manipulation tasks. During simulation, a proxy classifier $\mathbb{I}(\Vert {F_t} \Vert_2 > f_u)$ computes the force threshold $f_u$ breaches by the robot links $l$, leveraging the net contact force $F_t \in \mathbb{R}^{l \times 3}$ exposed by Isaac Gym. The full body contact awareness from this touch detector is crucial for the arm to leverage its multiple links (e.g. forearm and wrist) for manipulation, to supplement the partial, noisy, and time-varying point clouds. 

\textcolor{black}{Our simple reward structure listed below has three components: a) an occlusion clearance reward $r_h$, b) a smoothness reward $r_q$ on the joint velocities $\dot{q}$, and c) a safety bonus $r_{sm}$ to keep the arm clear of the power line safety margin,}
\begin{equation*}
r_h = \left[ \frac{1}{1 + h_t^2} \right]^2, \ r_q = -\sum_{j=1}^{6} \frac{\dot{q}_{tj}^2}{100}, \  r_{sm} = 0.4 \cdot \mathbb{I}_{\delta_{sm} = 0}.
\end{equation*}
\textcolor{black}{
The clearance reward $r_h \in [0, 1]$ is a monotonically decreasing function of the occlusion heuristic $h_t$. The $r_{sm} $ coefficient (0.4), is hand-tuned to make the sparse safety bonus noticeable without overshadowing clearance reward, whereas $r_q$ is a penalty term to encourage smoother trajectories.}

\begin{figure*}
\centering
\begin{tabular}{ccccc}
    \includegraphics[height=0.165\textwidth]{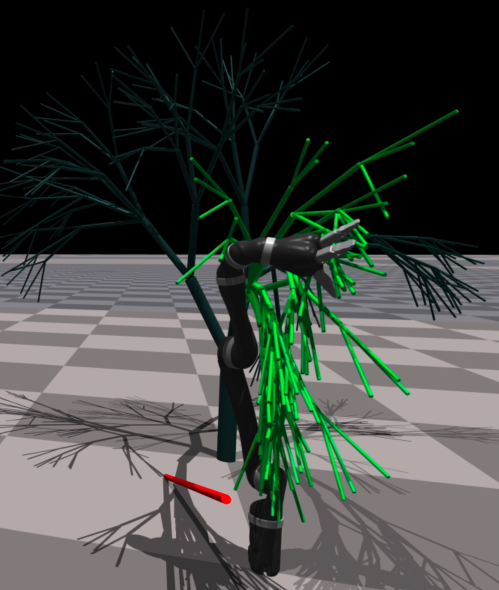} &
    \includegraphics[height=0.165\textwidth]{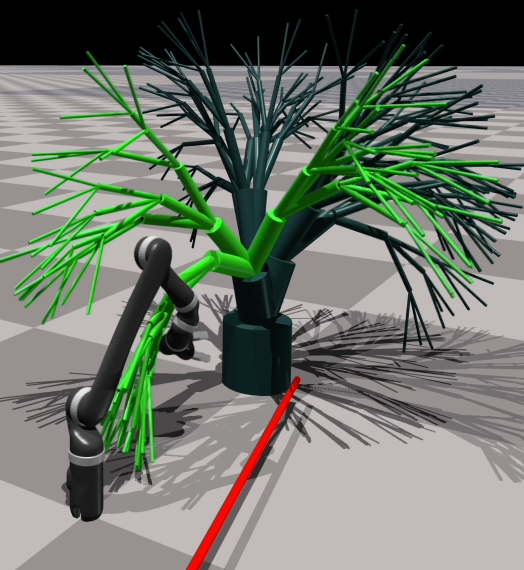} &
    \includegraphics[height=0.165\textwidth]{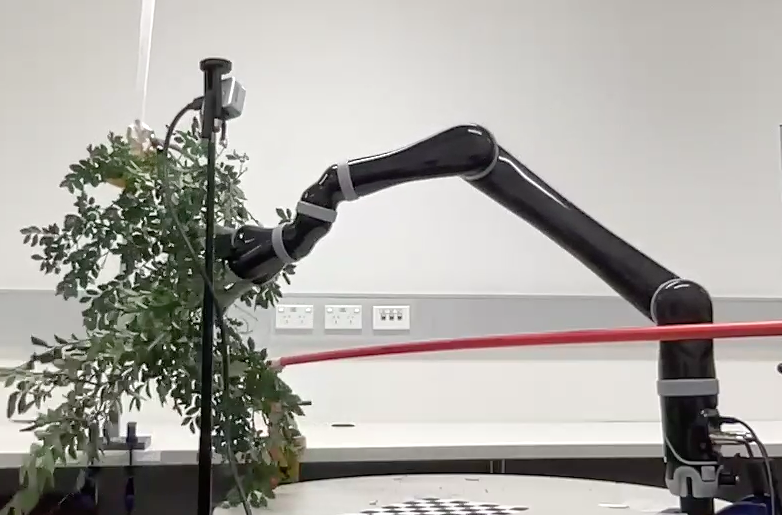} &
    \includegraphics[height=0.165\textwidth]{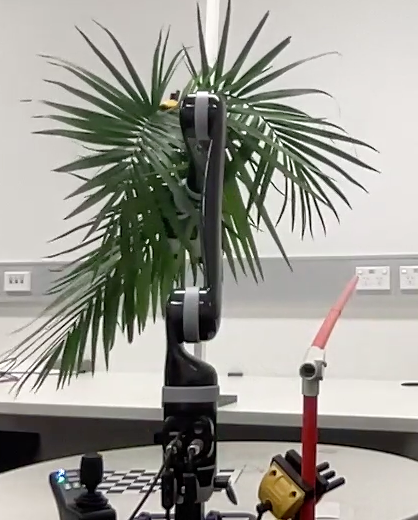} &
    \includegraphics[height=0.165\textwidth]{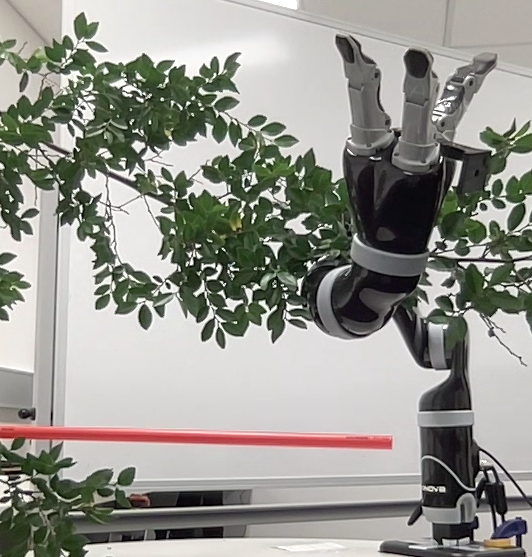} \\
    (a) & (b) & (c) & \textcolor{black}{(d)} & (e)
\end{tabular}

\caption{Terminal state poses: (a)(b): Simulation trajectory states showing the whole arm being utilised to shield the power line. (c)-(e): Similar real strategies using various arm links for clearance, \textcolor{black}{executed on branches of different tree species.}}
\label{fig:final-states}
\end{figure*}

\section{EXPERIMENTAL SETUP}

\subsection{Simulation Setting}
\label{sec:simulation_experimental_setting}
\textcolor{black}{For power line clearance (Fig. \ref {fig:problem-description}b), we run 6144 parallel Gym environments, each simulating a tree with deformable branches (in green), a rigid power line (in red), and a Kinova arm actuated via velocity control of its six joints, omitting the end-effector.} Each tree is formed by drawing from a Gaussian distribution ($\sigma= 0.1$) representing the L-system morphology parameters to ensure high diversity in topology but sufficient physical feasibility, in addition to arbitrary trunk rotation to expose the agent to varying branch occlusion patterns. 
Furthermore, the power line location is randomised, but constrained to the reach region of the gripper.
Finally, we perform all experiments with an average $h_t \in [0, 1]$ of at least $0.7$ across all environments, noting that occlusion score can increase as well from the arm pushing more branches closer to the power line.

A critical challenge within this framework is the contact explosion resulting from the combinatorial interactions among the multi-link arm, multi-branch tree, and the power line, exacerbated by a large number of training environments. This issue, common in contact-rich parallel simulations, can quickly overwhelm the simulation constraint solvers, triggering object inter-penetrations and CUDA memory overflows. 
We take a shortcut to contact reduction at the cost of a marginal drop in test performance. Specifically, during training, we allow the branches (but not the robot) to penetrate the power line through collision masking, thereby removing one key set of interactions. 
In contrast, during simulation tests, all contacts are in place to accurately reflect real-world, but the environment count is set low.
This approach, along with careful tuning of the Temporal Gauss-Seidel (TGS) contact solver parameters, enables us to run large-scale training without compromising representational accuracy.

\begin{table*}[t]
\centering
\caption{\textcolor{black}{Baseline Comparison \& Feature Ablation}}
\begin{tabular}{clcccccccc}
\toprule
\toprule
\textbf{ } & \textbf{Description} & \textbf{$d_{\text{max}}$} &  \textbf{Train Rew} & \textbf{Trials} & \textbf{Test Rew} & \textbf{Test SR\%} & \textbf{Occ Drop\%} & \textbf{Touch\%} & \textbf{Steps in Succ} \\
\midrule

\color{black} (a) & \color{black} Baseline: Guided IK - Lift & \color{black} -  & \color{black} - & \color{black} 1536 & \color{black} -  & \color{black} 33.4 ± 2.7  & \color{black} 12.7 ± 1.4 & \color{black} 11.3 ± 0.4 & \color{black} 179.1 ± 13.5 \\
\color{black} (b) & \color{black} Baseline: Guided IK - Sweep & \color{black} -  & \color{black} - & \color{black} 1536 & \color{black} -  & \color{black} 41.1 ± 1.7  & \color{black} 17.2 ± 2.2 & \color{black} 11.0 ± 0.6 & \color{black} 193.5 ± 17.5 \\
\color{black} (c) & \color{black} Baseline: GNN Features & \color{black} 0.02  & \color{black} 1613.9 & \color{black} 1536 & \color{black} 1511.7 ± 21.0 & \color{black} 55.5 ± 2.9 & \color{black} 33.6 ± 1.7 & \color{black} 10.5 ± 0.1 & \color{black} 291.4 ± 14.6 \\
\midrule
(d) & Proprioceptive + Heuristic:$h_t$ & -  & 1495.6 & 1536 & 1413.6 ± 28.7 & 36.5 ± 2.3 & 18.0 ± 1.3 & 2.8 ± 0.3 & 237.7 ± 19.2 \\
(e) & (d) + Local PC Features & 0.02 & 1531.7 & 1536 & 1486.5 ± 49.1 & 43.8 ± 4.5 & 25.2 ± 3.6 & 2.1 ± 0.4 & 258.3 ± 32.1 \\
(f) & (e) + Touch Feature & 0.02  & 1606.6 & 1536 & 1506.1 ± 28.5 & 46.1 ± 2.6 & 27.9 ± 4.8 & 2.6 ± 0.8 & 268.9 ± 17.7 \\
(g) & (e) + KME Features & 0.02 & 1739.1 & 1536 & 1630.2 ± 31.6 & 55.7 ± 1.1 & 40.1 ± 0.4 & 1.9 ± 0.1 & 348.2 ± 22.7 \\
(h) & All Features (RBF) & 0.02 & 1704.3 & 1536 &  1655.9 ± 26.4 & 63.4 ± 2.8 & 46.3 ± 3.3 & 3.8 ± 0.3 & 363.4 ± 20.6 \\
\midrule
\color{black} (i) & \color{black} All Features (Laplace) & \color{black} 0.02 & \color{black} 1592.8 & \color{black} 1536 & \color{black} 1559.0 ± 35.0 & \color{black} 52.5 ± 1.8 & \color{black} 36.0 ± 2.4 & \color{black} 4.0 ± 0.2 & \color{black} 315.9 ± 21.0 \\
\color{black} (j) & \color{black} All Features (Matérn-3/2) & \color{black} 0.02 & \color{black} 1662.5 & \color{black} 1536 & \color{black} 1620.0 ± 25.5 & \color{black} 59.8 ± 1.3 & \color{black} 43.5 ± 1.5 & \color{black} 3.2 ± 0.2 & \color{black} 362.9 ± 16.7 \\

\midrule
(k) & {\textbf{Final Model}} & 0.005 & \textbf{1825.1} & \textbf{1536} & \textbf{1763.6 ± 21.5} & \textbf{68.4 ± 2.1} & \textbf{53.7 ± 2.2} & \textbf{3.3 ± 0.4} & \textbf{429.8 ± 25.6} \\
\bottomrule
\bottomrule
\end{tabular}
\label{tab:ablation_study}
\end{table*}

\subsection{Hardware Design \& Sim-to-Real}
\label{sec:hardware_and_sim_to_real}
In this work, we aimed to train a policy in simulation, running on an NVIDIA RTX 4090, and transfer it to real without a policy adaptation phase. During inference, the agent interacts with our real Kinova 6-DOF Jaco 2 through the Kinova ROS API and a custom REST interface, running on a separate low-grade workstation at a 60Hz control rate. Unlike prior works focused on artificial plants with soft leafy resistance, our test branches are randomly chosen from real trees, ensuring sufficient stiffness to resist clearance. A single fixed Intel RealSense D405 camera, operating at 30 fps, provides the RGB-D ($480 \times 640$) observation of the workspace, illustrated in Fig. \ref {fig:final-states}:(c)-(e). The segmented real branch point clouds are constructed from the RGB-D camera images; in contrast, simulation tree points are randomly sampled from the coarse-grained branch segment surfaces. The real robot point cloud is built once from meshes, just as in the simulation, and updated at each step with the 3D coordinate transformation tree from ROS. 

Furthermore, to address the significant sim-to-real discrepancies in robot parameters, contact dynamics, and gravity compensation, exacerbated by velocity control, we leverage Segmented Steady State Error Control (SSED)\cite{jacobgentle}. This scheme alternates between applying an action $a_t \sim \pi_w(\cdot | s_t)$ to the desired state $s^d_t$, instead of the current, but periodically synchronising $s^d_t$ to $s_t$, in effect, rejecting short-term steady-state errors and clearing long-term accumulated offsets.

\subsection{Real Vision Pipeline}
\label{sec:real_vision_pipeline}
This section outlines the key transformation steps required to convert the real-time streaming images into RL features. First, we use Grounding DINO \cite{liu2023grounding} to locate the objects in the scene with simple text prompts. The resulting bounding boxes are passed to a Segment Anything SAM-HQ \cite{ke2023segment} model to generate object masks. The masks are then de-projected to the robot frame, leveraging the depth channel input. Crucially, manipulators operating in velocity drive mode have a control rate lower bound, 60Hz in our case, required to maintain velocity. Current state-of-the-art grounding and segmentation tools are unable to achieve this without compromising the mask quality. Therefore, we use a video segmentation framework Cutie \cite{cheng2024putting}, to propagate the mask across time-steps, occasionally replacing it with updated masks to prevent quality loss. 

\section{EXPERIMENTS AND RESULTS}

\begin{figure}
    \centering
    \includegraphics[width=0.49\textwidth]{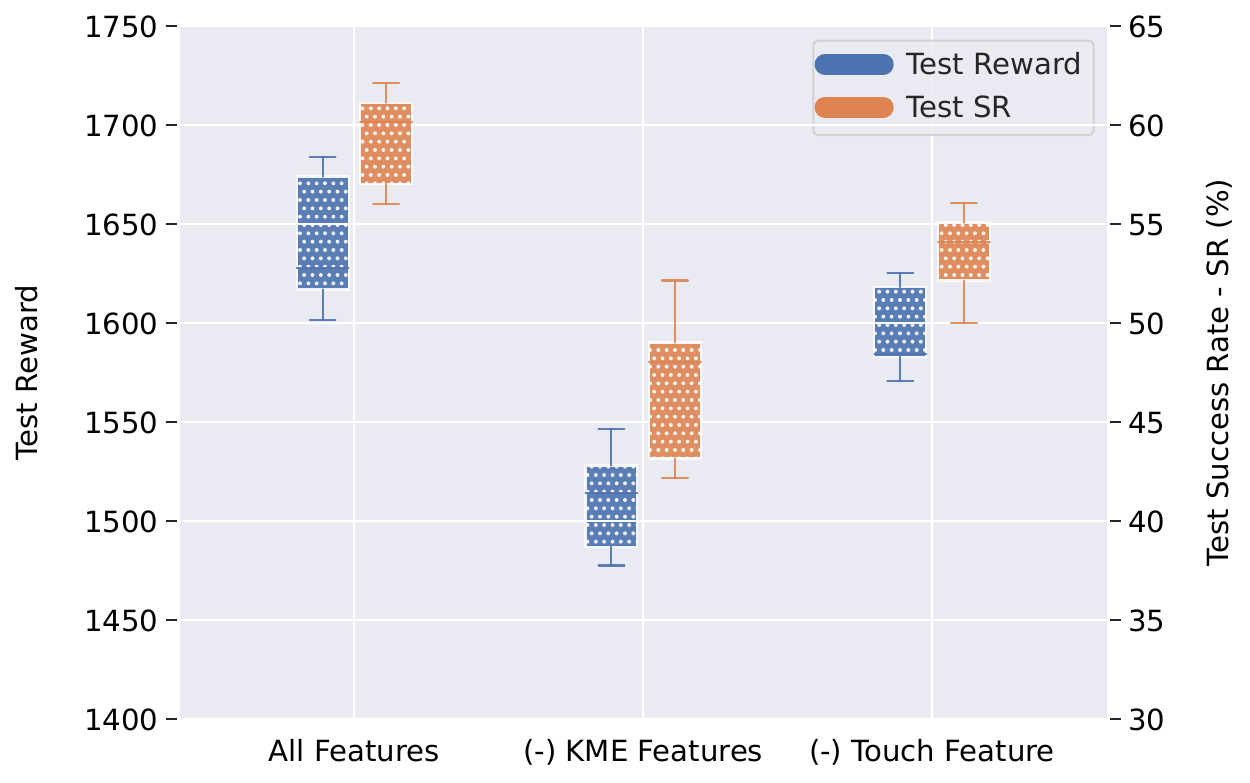}
    \caption{Ablation showing the relevance of key feature groups in our multi-modal policy. (-) indicates the removal of the single specified group. Each data point is a trained policy with a varying noise level. Boxes indicate median and IQR.}
    \label{fig:07_basic_ablation}
    \vspace*{-5mm}
\end{figure}

\textbf{Experiment 1}: \textcolor{black} {First, we establish quantitative non-RL baselines by executing hand-tuned, multi-step, inverse kinematics (IK) strategies in simulation.  The results are shown in Table \ref{tab:ablation_study}, rows (a)-(b). In the first case (Guided IK - Lift), we drive the end-effector to a location just above the power line mid-point (where the occlusion is maximum, on average), and then lift the arm up, to push the branches away. In the next case (Guided IK - Sweep), the arm first aligns the gripper to the axis connecting the robot base and tree trunk, then executes a sweeping motion to the left, if  power line is on the left of the robot, or to the right otherwise, clearing branches on the way.}

\textcolor{black}{As a second baseline, in Table \ref{tab:ablation_study}: (c), we leverage a GNN model \cite {kim2024towards}, where nodes represent branch fork positions and edges denote parent–child cylindrical links. However, unlike their graph-to-graph contact policy, which requires a predefined target deformation state (unknown in our context), we feed all node, edge, and global attributes to an RL policy, using only the representation component of the graph.}

\textbf{Experiment 2}: \ Second, we perform a simulation feature ablation study, \textcolor{black} {with RL policies}, to justify our choices. For each combination, we train independent policies by injecting varying levels of Gaussian Noise $\mathcal{N}(0, \sigma^2)$ into the point clouds with the $3\sigma$ rule, i.e., $\sigma = d_{\text{max}}/3$, $d_{\text{max}}$ being the maximum deviation representative of the camera depth inaccuracy. The feature ablation results are shown in Fig. \ref {fig:07_basic_ablation} and Table \ref{tab:ablation_study}: (d)-(h). The box plot data points are policies constructed with $d_{\text{max}}$ levels ranging from $0.005$ to $0.04$, displaying the distribution of test rewards and success rates on a single unseen test set. On the other hand, each row in Table \ref{tab:ablation_study} is an individual policy tested on multiple test sets, listing a variety of additional evaluation metrics, with rows (e)-(j) showing the results applying a median noise $d_{\text{max}}=0.02$, representative of the ablation study. By contrast, row (k) shows our best model with a lower $d_{\text{max}}=0.005$, the depth inaccuracy of the D405 camera used for real tests.
\textcolor{black}{
Furthermore, to justify our choice of RBF kernel,
we ablate against two alternatives: the Laplace kernel 
$k_{\mathrm{L}}(x, x') = \exp\!\left(-\|x - x'\| / \gamma\right)$
and Matérn-3/2 kernel 
$k_{\mathrm{M}}(x, x') = \left(1 + \sqrt{3}\|x - x'\| / \gamma\right)\exp\!\left(-\sqrt{3}\|x - x'\| / \gamma\right)$,
where, $\gamma > 0$ is a length-scale hyper-parameter. The kernel ablation results are in Table \ref{tab:ablation_study}: (i)-(j).
}

\begin{table*}[ht]
\centering
\begin{minipage}[t]{0.7\textwidth}
\centering
\caption{Representational Efficiency}
\label{tab:efficiency_comparison}
\begin{tabular}{lccccc}
\toprule
\toprule
  &  $\bm{GF_{t}^{(1024)}}$ (secs) & $\bm{GF_{t}^{(1)}}$ (secs) & \textbf{Train Time} (mins) & \textbf{Train Rew}  & \textbf{Max Envs} \\
\midrule
PointNet++ & 21.29  & 0.7706  & 1487.2  & 1345.6  & 1024\\
Point2Vec & 0.204  & 0.0454  & 75.2  & 1379.6  & 1024\\
\midrule
\textbf{KME} & \textbf{0.0012}  & \textbf{0.0012}  & \textbf{27.5}  & \textbf{1449.9}  & \textbf{6144}\\
\bottomrule
\bottomrule
\end{tabular}

\end{minipage}
\hfill
\begin{minipage}[t]{0.28\textwidth}
\centering
\caption{Real Results}
\label{tab:real_results}
\begin{tabular}{lcc}
\toprule
\toprule
\textbf{Branches} & \textbf{Attempts} & \textbf{Test SR} \\
\midrule
Branch 1 & 9  & 44.4\% \\
Branch 2 & 26 & 42.3\% \\
\textcolor{black}{Branch 3} & \textcolor{black}{18} & \textcolor{black}{55.5\%} \\
\bottomrule
\bottomrule
\end{tabular}
\end{minipage}
\vspace*{-4mm}
\end{table*}
\textbf{Experiments 1 \& 2 Results}: \textcolor{black}{The Table \ref{tab:ablation_study} result metrics are described as follows: Applicable only to RL, Training Reward (\textbf{Train Rew}) and Test Reward (\textbf{Test Rew}) assess the agent's performance during training and on unseen test scenarios.} Test Success Rate (\textbf{Test SR \%}) quantifies the percentage of test environments (\textbf{Trials}) where all occlusions were removed from the power line. Given that the branch collection will continue to resist clearance even after success is reached, we choose a conservative Test SR, considering a task as a success only if the line is occlusion-free for at least 10 consecutive steps. Occlusion reduction (\textbf{Occ Drop \%}) is the percentage drop in heuristic ($h_t$) from the start to end of test episodes averaged across all environments, while the (\textbf{Touch \%}) reflects the frequency of undesirable instances where the arm breaches the power line safety margin $\delta_{sm}$. The metric (\textbf{Steps in Succ}) measures the efficiency of successful trajectories; as an example, the final row (k) implies that, in a $1000$ step trajectory, the arm starting from the home position, traversed to the occlusion locations and moved the occluding branches away from the line in $1000 - 429.8 = 570.2$ steps, while the remaining $429.8$ steps were completely occlusion free for potential pruning, despite continued cluster resistance. All metrics except the power line touch (Touch \%) are higher, the better.

\textcolor{black}{From Table \ref{tab:ablation_study}: (a)-(c), the human-specified trajectories consistently underperformed our learning strategy. Our approach also outperformed the GNN simulation baseline; moreover, the graph method is less transferable to real, as it relies on true branch node positions that must be tracked across time-steps. Predictably, both baselines resulted in significantly higher power line contacts due to the absence of a touch indicator or a contact penalty. The improved Steps in Succ for RL policies indicate that, following successful de-occlusion, the RL agent attempts to stabilise the scene by holding the branches steady, unlike guided swipes. }

\textcolor{black}{From the feature ablation study, the plot in Fig. \ref {fig:07_basic_ablation} demonstrates that both the KME feature group capturing the global point cloud attributes and the contact detection classifier form essential constituents of our multi-modal policy, and removal of either would hurt performance. Table. \ref{tab:ablation_study}: (d)-(h) demonstrates that despite the stochasticity of individual policies and high noise injection, progressive addition of feature groups induces improvement across all evaluation metrics.} To highlight, we report all success rates without distinguishing between feasible and infeasible tasks, i.e., all occlusion patterns are randomly generated and contain scenarios that cannot be solved by a single arm. In some unsolvable failure instances, the arm cannot generate sufficient torques (due to joint limits) to overcome the collective branch resistance; in others, multiple disconnected branches can occlude the line at locations not closely spaced, only one of which can be removed at a time.

\textbf{Experiment 3}: Third, we compare the representational efficiency of our approach to other point cloud feature extractors, namely, PointNet++ \cite{qi2017pointnet++} and the Point2Vec \cite{zeid2023point2vec}. Specifically, we replace the four global point features from Table \ref{tab:rl_obs_features}, i.e., $\overline{\mu_{\mathbb{P}}}[P^{wbr}]$, $\overline{\mu_{\mathbb{P}}}[P^{zbr}]$ ,  $\overline{\mu_{\mathbb{P}}}[P^{clr}]$ and $\overline{\mu_{\mathbb{P}}}[P^{rob}]$ with embeddings from PointNet++ and Point2Vec. In each case, we use pre-trained checkpoints, down-sample the input, and employ PCA post-feature extraction for consistency with our approach. We train each set for 250 epochs with 1024 environments, tabulating the time taken (\textbf{Train Time}) and the accumulated reward (\textbf{Train Rew}). In addition, we compute ${GF_{t}^{(E)}}$ denoting the average time taken to build the 4x global point features only for a single RL time-step for $E$ environments. \bm{$GF_{t}^{(1024)}$} is indicative of the representation efficiency while training on 1024 environments and \bm{$GF_{t}^{(1)}$} represents the inference performance for one environment. The results are in Table \ref {tab:efficiency_comparison}, where lower is better for the time metrics, such as, \bm{$GF_{t}^{(1024)}$}, \bm{$GF_{t}^{(1)}$}, and \textbf{Train Time} while higher values are better for others.


Comparing the time taken for a single time-step and for the overall training, our kernel-based approach shows orders of magnitude improvement, with similar training rewards. For the PointNet++ and Point2Vec, the maximum possible environments (\textbf{Max Envs}) we could spin up in our hardware was 1024, an indication of the GPU memory constraint from heavy pre-trained checkpoints. Notice that our final KME-based model training, row (k) from Table \ref{tab:ablation_study}, which uses 6x environments and 5x iterations compared to this experiment, takes just 4.5 hours to complete on an RTX 4090. Furthermore, \bm{$GF_{t}^{(1)}$} values indicate that both PointNet++ and Point2Vec cannot meet the 60Hz real-time inference frequency requirement, justifying our approach. However, we acknowledge that improved efficiency may be at the cost of expressivity and that advanced baseline versions or quantisation techniques can reduce this time difference.


\textbf{Experiment 4}: Finally, we go beyond simulation validation to evaluate our approach in a real-world laboratory setting. We test \textcolor{black}{three} branches from different species, modifying either the branch orientation or the power line position in each experiment. Furthermore, in each test, we start with an $h_t$ of at least $0.8$, and consider both partial occlusion clearance and any touch to the power line as failure instances. 

The results (Table \ref{tab:real_results}) and the corresponding videos (supplement \& Fig. \ref{fig:final-states}) demonstrate that the simulation-trained policy can adapt zero-shot to arbitrary branch structures with multiple forking patterns despite the presence of point cloud noise and leaf-induced clutter unseen in training. The trajectories exhibit novel characteristics, effectively using the forearm, upper and lower wrists, gripper, and the unactuated end-effector. We observe self-reconfiguring strategies to prop up the occluding branches, while in others, the arm inserts itself between the branches and the line to separate them. Further, these results validate our RL approach in discovering efficient strategies and adapting to complex scenarios, well beyond what humans can demonstrate.

\section{LIMITATIONS AND FUTURE WORK}

A few directions for future work to reduce the sim-to-real gap and improve the success rates are as follows. First, the quality of perception can be enhanced with multiple camera views or with partial-to-complete point cloud models for surface reconstruction. 
Second, to introduce additional real-world complexities during simulation, integrating with system identification methods \cite{jacob2024learning} that infers parameters governing branch dynamics, may be necessary.
Third, external whole-arm tactile sensors \cite{huang20243d} could be leveraged instead of our proprioceptive-only model.
\textcolor{black}{Finally, our approach could extend beyond agriculture; for instance, to clear deformable cable bundles obstructing inspection cameras in data-centres, reposition flexible ducting during aircraft maintenance, or manipulate soft tissue clusters under occlusion in robotic surgery using whole-arm contact.}






\bibliographystyle{IEEEtran}
\bibliography{main} 

@article{antonova2022bayesian,
  title={A bayesian treatment of real-to-sim for deformable object manipulation},
  author={Antonova, Rika and Yang, Jingyun and Sundaresan, Priya and Fox, Dieter and Ramos, Fabio and Bohg, Jeannette},
  journal={IEEE Robotics and Automation Letters},
  volume={7},
  number={3},
  pages={5819--5826},
  year={2022},
  publisher={IEEE}
}

@article{yao2024safe,
  title={Safe Leaf Manipulation for Accurate Shape and Pose Estimation of Occluded Fruits},
  author={Yao, Shaoxiong and Pan, Sicong and Bennewitz, Maren and Hauser, Kris},
  journal={arXiv preprint arXiv:2409.17389},
  year={2024}
}

@inproceedings{jacobgentle,
  title={Gentle Manipulation of Tree Branches: A Contact-Aware Policy Learning Approach},
  author={Jacob, Jay and Cai, Shizhe and Borges, Paulo Vinicius Koerich and Bandyopadhyay, Tirthankar and Ramos, Fabio},
  booktitle={8th Annual Conference on Robot Learning}
}

@article{liu2023grounding,
  title={Grounding dino: Marrying dino with grounded pre-training for open-set object detection},
  author={Liu, Shilong and Zeng, Zhaoyang and Ren, Tianhe and Li, Feng and Zhang, Hao and Yang, Jie and Li, Chunyuan and Yang, Jianwei and Su, Hang and Zhu, Jun and others},
  journal={arXiv preprint arXiv:2303.05499},
  year={2023}
}

@article{ke2023segment,
  title={Segment anything in high quality},
  author={Ke, Lei and Ye, Mingqiao and Danelljan, Martin and Tai, Yu-Wing and Tang, Chi-Keung and Yu, Fisher and others},
  journal={Advances in Neural Information Processing Systems},
  volume={36},
  pages={29914--29934},
  year={2023}
}

@inproceedings{cheng2024putting,
  title={Putting the object back into video object segmentation},
  author={Cheng, Ho Kei and Oh, Seoung Wug and Price, Brian and Lee, Joon-Young and Schwing, Alexander},
  booktitle={Proceedings of the IEEE/CVF Conference on Computer Vision and Pattern Recognition},
  pages={3151--3161},
  year={2024}
}

@article{guggenmoos2003effects,
  title={Effects of tree mortality on power line security},
  author={Guggenmoos, Siegfried},
  journal={Arboriculture \& Urban Forestry (AUF)},
  volume={29},
  number={4},
  pages={181--196},
  year={2003},
  publisher={Arboriculture \& Urban Forestry (AUF)}
}

@article{arriola2020modeling,
  title={Modeling of deformable objects for robotic manipulation: A tutorial and review},
  author={Arriola-Rios, Veronica E and Guler, Puren and Ficuciello, Fanny and Kragic, Danica and Siciliano, Bruno and Wyatt, Jeremy L},
  journal={Frontiers in Robotics and AI},
  volume={7},
  pages={82},
  year={2020},
  publisher={Frontiers Media SA}
}

@article{yin2021modeling,
  title={Modeling, learning, perception, and control methods for deformable object manipulation},
  author={Yin, Hang and Varava, Anastasia and Kragic, Danica},
  journal={Science Robotics},
  volume={6},
  number={54},
  pages={eabd8803},
  year={2021},
  publisher={American Association for the Advancement of Science}
}

@article{lowe2022identification,
  title={The identification and management of hazard trees to mitigate bushfire risk},
  author={Lowe, Thomas and Lichman, Serge and Pinskier, Josh and Sun, Changming and Dunstall, Simon},
  year={2022}
}

@inproceedings{siebert2014survey,
  title={A survey of applied robotics for tree pruning near overhead power lines},
  author={Siebert, Luciano C and Toledo, Luiz FRB and Block, Pedro AB and Bahlke, Diogo B and Roncolatto, Ronaldo A and Cerqueira, Dailton P},
  booktitle={Proceedings of the 2014 3rd International Conference on Applied Robotics for the Power Industry},
  pages={1--5},
  year={2014},
  organization={IEEE}
}

@article{lin2022planning,
  title={Planning with spatial-temporal abstraction from point clouds for deformable object manipulation},
  author={Lin, Xingyu and Qi, Carl and Zhang, Yunchu and Huang, Zhiao and Fragkiadaki, Katerina and Li, Yunzhu and Gan, Chuang and Held, David},
  journal={arXiv preprint arXiv:2210.15751},
  year={2022}
}

@inproceedings{seita2023toolflownet,
  title={Toolflownet: Robotic manipulation with tools via predicting tool flow from point clouds},
  author={Seita, Daniel and Wang, Yufei and Shetty, Sarthak J and Li, Edward Yao and Erickson, Zackory and Held, David},
  booktitle={Conference on Robot Learning},
  pages={1038--1049},
  year={2023},
  organization={PMLR}
}

@article{zhou2022intelligent,
  title={Intelligent robots for fruit harvesting: Recent developments and future challenges},
  author={Zhou, Hongyu and Wang, Xing and Au, Wesley and Kang, Hanwen and Chen, Chao},
  journal={Precision Agriculture},
  volume={23},
  number={5},
  pages={1856--1907},
  year={2022},
  publisher={Springer}
}

@article{maillard2011selecting,
  title={Selecting the state-representation in reinforcement learning},
  author={Maillard, Odalric-Ambrym and Ryabko, Daniil and Munos, R{\'e}mi},
  journal={Advances in Neural Information Processing Systems},
  volume={24},
  year={2011}
}

@article{grannen2020untangling,
  title={Untangling dense knots by learning task-relevant keypoints},
  author={Grannen, Jennifer and Sundaresan, Priya and Thananjeyan, Brijen and Ichnowski, Jeffrey and Balakrishna, Ashwin and Hwang, Minho and Viswanath, Vainavi and Laskey, Michael and Gonzalez, Joseph E and Goldberg, Ken},
  journal={arXiv preprint arXiv:2011.04999},
  year={2020}
}

@article{deng2024general,
  title={General-purpose Clothes Manipulation with Semantic Keypoints},
  author={Deng, Yuhong and Hsu, David},
  journal={arXiv preprint arXiv:2408.08160},
  year={2024}
}

@article{jain2013reaching,
  title={Reaching in clutter with whole-arm tactile sensing},
  author={Jain, Advait and Killpack, Marc D and Edsinger, Aaron and Kemp, Charles C},
  journal={The International Journal of Robotics Research},
  volume={32},
  number={4},
  pages={458--482},
  year={2013},
  publisher={SAGE Publications Sage UK: London, England}
}

@article{huang20243d,
  title={3D-ViTac: Learning Fine-Grained Manipulation with Visuo-Tactile Sensing},
  author={Huang, Binghao and Wang, Yixuan and Yang, Xinyi and Luo, Yiyue and Li, Yunzhu},
  journal={arXiv preprint arXiv:2410.24091},
  year={2024}
}

@inbook{prusinkiewicz2012algorithmic,
  author={Prusinkiewicz, Przemyslaw and Lindenmayer, Aristid},
  title={The algorithmic beauty of plants},
  chapter={2},
  publisher={Springer Science \& Business Media},
  year={2012},
  pages={58--61}
}

@article{jacob2024learning,
  title={Learning to simulate tree-branch dynamics for manipulation},
  author={Jacob, Jayadeep and Bandyopadhyay, Tirthankar and Williams, Jason and Borges, Paulo and Ramos, Fabio},
  journal={IEEE Robotics and Automation Letters},
  volume={9},
  number={2},
  pages={1748--1755},
  year={2024},
  publisher={IEEE}
}

@article{makoviychuk2021isaac,
  title={Isaac gym: High performance gpu-based physics simulation for robot learning},
  author={Makoviychuk, Viktor and Wawrzyniak, Lukasz and Guo, Yunrong and Lu, Michelle and Storey, Kier and Macklin, Miles and Hoeller, David and Rudin, Nikita and Allshire, Arthur and Handa, Ankur and others},
  journal={arXiv preprint arXiv:2108.10470},
  year={2021}
}

@article{muandet2017kernel,
  title={Kernel mean embedding of distributions: A review and beyond},
  author={Muandet, Krikamol and Fukumizu, Kenji and Sriperumbudur, Bharath and Sch{\"o}lkopf, Bernhard and others},
  journal={Foundations and Trends{\textregistered} in Machine Learning},
  volume={10},
  number={1-2},
  pages={1--141},
  year={2017},
  publisher={Now Publishers, Inc.}
}

@article{rahimi2007random,
  title={Random features for large-scale kernel machines},
  author={Rahimi, Ali and Recht, Benjamin},
  journal={Advances in neural information processing systems},
  volume={20},
  year={2007}
}

@article{bai2025retrieval,
  title={Retrieval dexterity: Efficient object retrieval in clutters with dexterous hand},
  author={Bai, Fengshuo and Li, Yu and Chu, Jie and Chou, Tawei and Zhu, Runchuan and Wen, Ying and Yang, Yaodong and Chen, Yuanpei},
  journal={arXiv preprint arXiv:2502.18423},
  year={2025}
}

@article{subedi2025find,
  title={Find the Fruit: Designing a Zero-Shot Sim2Real Deep RL Planner for Occlusion Aware Plant Manipulation},
  author={Subedi, Nitesh and Yang, Hsin-Jung and Jha, Devesh K and Sarkar, Soumik},
  journal={arXiv preprint arXiv:2505.16547},
  year={2025}
}

@inproceedings{sundaresan2020learning,
  title={Learning rope manipulation policies using dense object descriptors trained on synthetic depth data},
  author={Sundaresan, Priya and Grannen, Jennifer and Thananjeyan, Brijen and Balakrishna, Ashwin and Laskey, Michael and Stone, Kevin and Gonzalez, Joseph E and Goldberg, Ken},
  booktitle={2020 IEEE International Conference on Robotics and Automation (ICRA)},
  pages={9411--9418},
  year={2020},
  organization={IEEE}
}

@article{tang2022track,
  title={Track deformable objects from point clouds with structure preserved registration},
  author={Tang, Te and Tomizuka, Masayoshi},
  journal={The International Journal of Robotics Research},
  volume={41},
  number={6},
  pages={599--614},
  year={2022},
  publisher={SAGE Publications Sage UK: London, England}
}

@article{liu2022frame,
  title={Frame mining: a free lunch for learning robotic manipulation from 3d point clouds},
  author={Liu, Minghua and Li, Xuanlin and Ling, Zhan and Li, Yangyan and Su, Hao},
  journal={arXiv preprint arXiv:2210.07442},
  year={2022}
}

@article{ma2021learning,
  title={Learning latent graph dynamics for deformable object manipulation},
  author={Ma, Xiao and Hsu, David and Lee, Wee Sun},
  journal={arXiv preprint arXiv:2104.12149},
  volume={2},
  year={2021}
}

@article{zhang2023push,
  title={Push past green: Learning to look behind plant foliage by moving it},
  author={Zhang, Xiaoyu and Gupta, Saurabh},
  journal={arXiv preprint arXiv:2307.03175},
  year={2023}
}

@article{yamamoto2014development,
  title={Development of a stationary robotic strawberry harvester with a picking mechanism that approaches the target fruit from below},
  author={Yamamoto, Satoshi and Hayashi, Shigehiko and Yoshida, Hirotaka and Kobayashi, Ken},
  journal={Japan Agricultural Research Quarterly: JARQ},
  volume={48},
  number={3},
  pages={261--269},
  year={2014},
  publisher={Japan International Research Center for Agricultural Sciences}
}

@inproceedings{yandun2021reaching,
  title={Reaching pruning locations in a vine using a deep reinforcement learning policy},
  author={Yandun, Francisco and Parhar, Tanvir and Silwal, Abhisesh and Clifford, David and Yuan, Zhiqiang and Levine, Gabriella and Yaroshenko, Sergey and Kantor, George},
  booktitle={2021 IEEE International Conference on Robotics and Automation (ICRA)},
  pages={2400--2406},
  year={2021},
  organization={IEEE}
}

@article{schulman2017proximal,
  title={Proximal policy optimization algorithms},
  author={Schulman, John and Wolski, Filip and Dhariwal, Prafulla and Radford, Alec and Klimov, Oleg},
  journal={arXiv preprint arXiv:1707.06347},
  year={2017}
}

@article{qi2017pointnet++,
  title={Pointnet++: Deep hierarchical feature learning on point sets in a metric space},
  author={Qi, Charles Ruizhongtai and Yi, Li and Su, Hao and Guibas, Leonidas J},
  journal={Advances in neural information processing systems},
  volume={30},
  year={2017}
}

@inproceedings{zeid2023point2vec,
  title={Point2vec for self-supervised representation learning on point clouds},
  author={Zeid, Karim Abou and Schult, Jonas and Hermans, Alexander and Leibe, Bastian},
  booktitle={DAGM German Conference on Pattern Recognition},
  pages={131--146},
  year={2023},
  organization={Springer}
}

@inproceedings{kim2024towards,
  title={Towards robotic tree manipulation: Leveraging graph representations},
  author={Kim, Chung Hee and Lee, Moonyoung and Kroemer, Oliver and Kantor, George},
  booktitle={2024 IEEE International Conference on Robotics and Automation (ICRA)},
  pages={11884--11890},
  year={2024},
  organization={IEEE}
}







\end{document}